\documentclass[11pt,a4paper]{article}
\usepackage{swisstextkonvens2020}
\usepackage{times}
\usepackage{latexsym}

\usepackage{url}
\usepackage{natbib}

\usepackage{booktabs}
\usepackage{tabularx}

\usepackage{graphicx}
\usepackage[T1]{fontenc}
\usepackage[utf8]{inputenc}
\graphicspath{{figures/}}

\aclfinalcopy 

\newcommand{\bert}{\textsc{Bert}}
\newcommand{\mbert}{\mbox{\textsc{M}-\bert{}}}
\newcommand{\xstance}{\mbox{\textsc{x}-stance}}

\title{\xstance{}: \\A Multilingual Multi-Target Dataset for Stance Detection}

\author{Jannis Vamvas$^1$ \quad Rico Sennrich$^{1,2}$ \bigskip\\
  $^1$Department of Computational Linguistics, University of Zurich\\
  $^2$School of Informatics, University of Edinburgh \smallskip\\
  \texttt{\{vamvas,sennrich\}@cl.uzh.ch}
}
\date{}

\begin{document}
\maketitle
\begin{abstract}
    We extract a large-scale stance detection data\-set from comments written by candidates of elections in Switzerland.
    The dataset consists of German, French and Italian text, allowing for a cross-lingual evaluation of stance detection.
    It contains 67\,000~comments on more than 150~political issues~(\textit{targets}).
    Unlike stance detection models that have specific target issues, we use the dataset to train a single model on all the issues.
    To make learning across targets possible, we prepend to each instance a natural question that represents the target (e.g.~``Do~you support~X?'').
    Baseline results from multilingual \bert{} show that zero-shot cross-lingual and cross-target transfer of stance detection is moderately successful with this approach.

\end{abstract}

\section{Introduction}

\begin{figure*}
\centering
\def\svgwidth{\textwidth}
\small{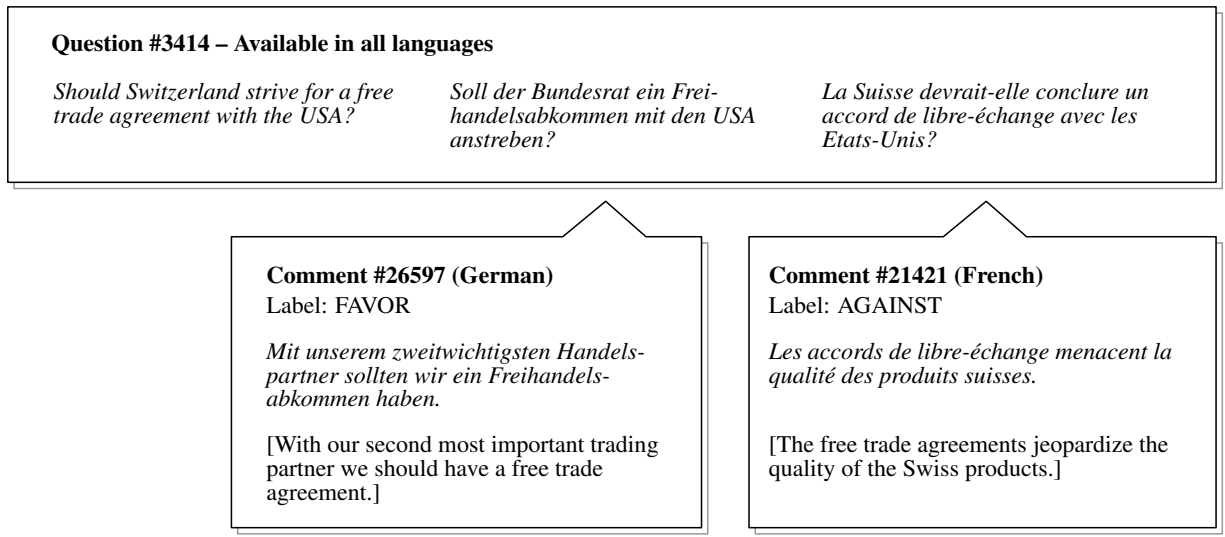}
\caption{
Example of a question and two answers in the \xstance{} dataset.
The answers were submitted by electoral candidates on a voting advice website.
The author of the German comment was in favor of the issue; the author of the French comment against.
Both authors use comments to explain their respective stance.
}
\label{example}
\end{figure*}

In recent years many datasets have been created for the task of automated stance detection, advancing natural language understanding systems for political science, opinion research and other application areas.
Typically, such benchmarks~\cite{mohammad-etal-2016-dataset} are composed of short pieces of text commenting on politicians or public issues and are manually annotated with their stance towards a \textit{target entity} (e.g. Climate Change, or Trump).
However, they are limited in scope on multiple levels~\cite{10.1145/3369026}.

First of all, it is questionable how well current stance detection methods perform in a cross-lingual setting, as the multilingual datasets available today are relatively small, and specific to a single target~\cite{taule2017overview, taule2018overview}.
Furthermore, specific models tend to be developed for each single target or pair of targets~\cite{sobhani-etal-2017-dataset}.
Concerns have been raised that cross-target performance is often considerably lower than fully supervised performance~\cite{10.1145/3369026}.

In this paper we propose a much larger dataset that combines multilinguality and a multitude of topics and targets.
\xstance{} comprises more than 150~questions about Swiss politics and more than 67k~answers given by candidates running for political office in Switzerland.
Questions are available in four languages: English, Swiss Standard German, French, and Italian.
The language of a comment depends on the candidate's region of origin.

We have extracted the data from the voting advice application \textit{Smartvote}.
Candidates respond to questions mainly in categorical form (yes / rather yes / rather no / no).
They can also submit a free-text comment to justify or explain their categorical answer.
An example is given in Figure~\ref{example}.

We transform the dataset into a stance detection task by interpreting the question as a natural-language representation of the target, and the commentary as the input to be classified.

The dataset is split into a multilingual training set and into several test sets to evaluate zero-shot cross-lingual and cross-target transfer.
To provide a baseline, we fine-tune a multilingual \bert{} model~\cite{devlin-etal-2019-bert} on \xstance{}.
We show that the baseline accuracy is comparable to previous  stance detection benchmarks while leaving ample room for improvement.
In addition, the model can generalize to a degree both cross-lingually and in a cross-target setting.

We have made the dataset and the code for reproducing the baseline models publicly available.\footnote{\url{http://doi.org/10.5281/zenodo.3831317}}

\section{Related Work}

\paragraph{Multilingual Stance Detection}

In the context of the IberEval shared tasks, two related multilingual datasets have been created ~\cite{taule2017overview, taule2018overview}.
Both are a collection of annotated Spanish and Catalan tweets.
Crucially, the tweets in both languages focus on the same issue (Catalan independence); given this fact they are the first truly multilingual stance detection datasets known to us.

With regard to the languages covered by \xstance{}, only monolingual datasets seem to be available.
For French, a collection of tweets on French presidential candidates has been annotated with stance~\cite{lai2020multilingual}.
Similarly, two datasets of Italian tweets on the occasion of the 2016 constitutional referendum have been created~\cite{lai2018stance, lai2020multilingual}.
With regard to German, a corpus of 270 sentences has been annotated with fine-grained stance and attitude information~\cite{clematide-etal-2012-mlsa}.
Furthermore, fine-grained stance detection has been qualitatively studied on a large corpus of Facebook posts~\cite{klenner-etal-2017-stance}.

\paragraph{Multi-Target Stance Detection}
The SemEval-2016 task on detecting stance in tweets~\cite{mohammad-etal-2016-semeval} offers data concerning multiple targets (Atheism, Climate Change, Feminism, Hillary Clinton, and Abortion).
In the supervised subtask A, participants tended to develop a \textit{target-specific} model for each of those targets.
In subtask B  cross-target transfer to the target ``Donald Trump'' was tested, for which no annotated training data were provided.
While this required the development of more universal models, their performance was generally much lower.

\citet{sobhani-etal-2017-dataset} introduced a \textit{multi-target} stance dataset which provides two targets per instance.
For example, a model designed in this framework is supposed to simultaneously classify a tweet with regard to Clinton and with regard to Trump.
While in theory the framework allows for more than two targets, it is still restricted to a finite and clearly defined set of targets.
It focuses on modeling the dependencies of multiple targets within the same text sample, while our approach focuses on learning stance detection from many samples with many different targets.

\paragraph{Representation Learning for Stance Detection}
In a target-specific setting, \citet{ghosh2019stance} perform a systematic evaluation of stance detection approaches.
They also evaluate \bert{}~\cite{devlin-etal-2019-bert} and find that it consistently outperforms previous approaches.

However, they only experiment with a single-segment encoding of the input, preventing cross-target transfer of the model.
\citet{augenstein-etal-2016-stance} propose a conditional encoding approach to encode both the target and the tweet as sequences.
They use a bidirectional LSTM to condition the encoding of the tweets on the encoding of the target, and then apply a nonlinear projection on the conditionally encoded tweet.
This allows them to train a model that can generalize to previously unseen targets.

\section{The \xstance{} Dataset}

\begin{table}[]
\begin{tabularx}{\columnwidth}{@{}Xrr@{}}
\toprule
\textbf{Topic}                   & \textbf{Questions} & \textbf{Answers} \\ \midrule
Digitisation                     & 2            & 1168           \\
Economy                          & 23           & 6899           \\
Education                        & 16           & 7639           \\
Finances                         & 15           & 3980           \\
Foreign Policy                   & 16           & 4393           \\
Immigration                      & 19           & 6270           \\
\makebox[0pt][l]{Infrastructure \& Environment} & 31           & 9590 \\
Security                         & 20           & 5193           \\
Society                          & 17           & 6275           \\
Welfare                          & 15           & 8508           \\
\textbf{Total (training topics)}    & \textbf{174} & \textbf{59 915} \\ [0.5ex] \midrule
Healthcare                       & 11           & 4711           \\
Political System                 & 9            & 2645           \\
\makebox[0pt][l]{\textbf{Total (held-out topics)}} & \textbf{20}  & \textbf{7356}  \\ \bottomrule
\end{tabularx}
\caption{\label{topics}
Number of questions and answers per topic.
}
\end{table}

\begin{table*}[]
\setlength{\tabcolsep}{22px}
\begin{tabularx}{\textwidth}{@{}Xl@{}rl@{}rl@{}r@{}}
\toprule
    & \multicolumn{2}{l}{\parbox{3.5cm}{\textbf{Intra-target} \\ (New answers to \\ known questions)}}
    & \multicolumn{2}{l}{\parbox{3.3cm}{\textbf{Cross-question} \\ (New questions \\ within known topics)}}
    & \multicolumn{2}{l}{\parbox{2.4cm}{\textbf{Cross-topic} \\ \phantom{()} \\ \phantom{()}}} \\ \midrule
    \textsc{de} & \begin{tabular}[c]{@{}l@{}}Train:\\ Test:\\ Valid:\end{tabular}
        & \begin{tabular}[c]{@{}r@{}}33 850\\ 2871\\ 3479\end{tabular}
        & Test: & 3143 & Test: & 5269 \\ \midrule
    \textsc{fr} & \begin{tabular}[c]{@{}l@{}}Train:\\ Test:\\ Valid:\end{tabular}
        & \begin{tabular}[c]{@{}r@{}}11 790\\ 1055\\ 1284\end{tabular}
        & Test: & 1170 & Test: & 1914 \\ \midrule
    \textsc{it} & Test: & 1173 & Test: & (110) & Test: & (173) \\ \bottomrule
\end{tabularx}
\caption{\label{split}
Number of answer instances in the training, validation and test sets.
The \textit{upper left corner} represents a multilingually supervised task, where training, validation and test data are from exactly the same domain.
The \textit{top-to-bottom axis} gives rise to a cross-lingual transfer task, where a model trained on German and French is evaluated on Italian answers to the same questions.
The \textit{left-to-right axis} represents a continuous shift of domain:
In the \textit{middle column}, the model is tested on previously unseen questions that belong to the same topics as seen during training.
In the \textit{right column} the model encounters unseen answers to unseen questions within an unseen topic.
The two test sets in parentheses are too small for a significant evaluation.
}
\end{table*}

\subsection{Task Definition}

The input provided by \xstance{} is two-fold: (A) a natural language question concerning a political issue; (B) a natural language commentary on a specific stance towards the question.

The label to be predicted is either `favor' or `against`.
This corresponds to a standard established by \citet{mohammad-etal-2016-dataset}.
However, \xstance{} differs from that dataset in that it lacks a `neither' class; all comments refer to either a `favor' or an `against` position.
The task posed by \xstance{} is thus a binary classification task.

As an evaluation metric we report the macro-average of the F1-score for `favor' and the F1-score for `against', similar to ~\citet{mohammad-etal-2016-semeval}.
We use this metric mainly to strengthen comparability with the previous benchmarks.

\subsection{Data Collection}

\paragraph{Provenance}
We downloaded the questions and answers via the \textit{Smartvote} API\footnote{\url{https://smartvote.ch}}. The downloaded data cover 175 communal, cantonal and national elections between 2011 and 2020.

All candidates in an election who participate in \textit{Smartvote} are asked the same set of questions, but depending on the locale they see translated versions of the questions.
They can answer each question with either `yes', `rather yes', `rather no', or `no'.
They can supplement each answer with a comment of at most 500 characters.

The questions asked on \textit{Smartvote} have been edited by a team of political scientists.
They are intended to cover a broad range of political issues relevant at the time of the election.
A detailed documentation of the design of \textit{Smartvote} and the editing process of the questions is provided by \citet{thurman2009three}.

\paragraph{Preprocessing}
We merged the two labels on each pole into a single label: `yes' and `rather yes' were combined into `favor'; `rather no', or `no' into `against`.
This improves the consistency of the data and the comparability to previous stance detection datasets.
We did not further preprocess the text of the comments.

\paragraph{Language Identification}
As the API does not provide the language of comments, we employed a language identifier to automatically annotate this information.
We used the \textit{langdetect} library~\cite{nakatani2010langdetect}.
For each responder we classified all the comments jointly, assuming that responders did not switch code during the answering of the questionnaire.

We applied the identifier in a two-step approach.
In the first run we allowed the identifier to output all 55 languages that it supports out of the box, plus Romansh, the fourth official language in Switzerland\footnote{Namely the Rumantsch Grischun variety; the language profile was created using resources from the Zurich Parallel Corpus Collection~\cite{GraenKewShaitarovaVolk2019} and the \textit{Quotidiana} corpus (\url{https://github.com/ProSvizraRumantscha/corpora}).}.
We found that no Romansh comments were detected and that all unexpected outputs were misclassifications of German, French or Italian comments.
We further concluded that little or no Swiss German comments are in the dataset; otherwise, some of them would have manifested themselves via misclassifications~(e.g.~as~Dutch).

In the second run, drawing from these conclusions, we restricted the identifier's set of choices to English, French, German and Italian.

\paragraph{Filtering}
We pre-filtered the questions and answers to improve the quality of the dataset.
To keep the domain of the data surveyable, we set a focus on national-level questions.
Therefore, all questions and corresponding answers pertaining to national elections were included.

In the context of communal and cantonal elections, candidates have answered both local questions and a subset of the national questions.
Of those elections, we only considered answers to the questions that also had been asked in a national election.
They were only used to augment the training set while the validation and test sets were restricted to answers from national elections.

We discarded the fewer than 20 comments classified as English.
Furthermore, we discarded instances that met any of the following conditions:
\begin{itemize}\itemsep0em
    \item Question is not a closed question or does not address a clearly defined political issue.
    \item No comment was submitted by the candidate or the comment is shorter than 50 characters.
    \item Comment starts with ``but'' or a similar indicator that the comment is not self-contained.
    \item Comment contains a URL.
\end{itemize}

\noindent In total, a fifth of the comments were filtered out.

\paragraph{Topics}

The questions have been organized by the \textit{Smartvote} editors into categories (such as~``Economy'').
We further consolidated the pre-defined categories into 12 broad topics (Table~\ref{topics}).

\paragraph{Compliance}
The dataset is shared under a CC BY-NC 4.0 license.
Copyright remains with www.smartvote.ch.

Given the sensitive nature of the data, we increase the anonymity of the data by hashing the respondents' IDs.
No personal attributes of the respondents are included in the dataset.
We provide a data statement~\cite{doi:10.1162/tacl-a-00041} in Appendix~\ref{appendix:data-statement}.

\subsection{Data Split}

\begin{figure*}
\centering
\def\svgwidth{\textwidth}
\small{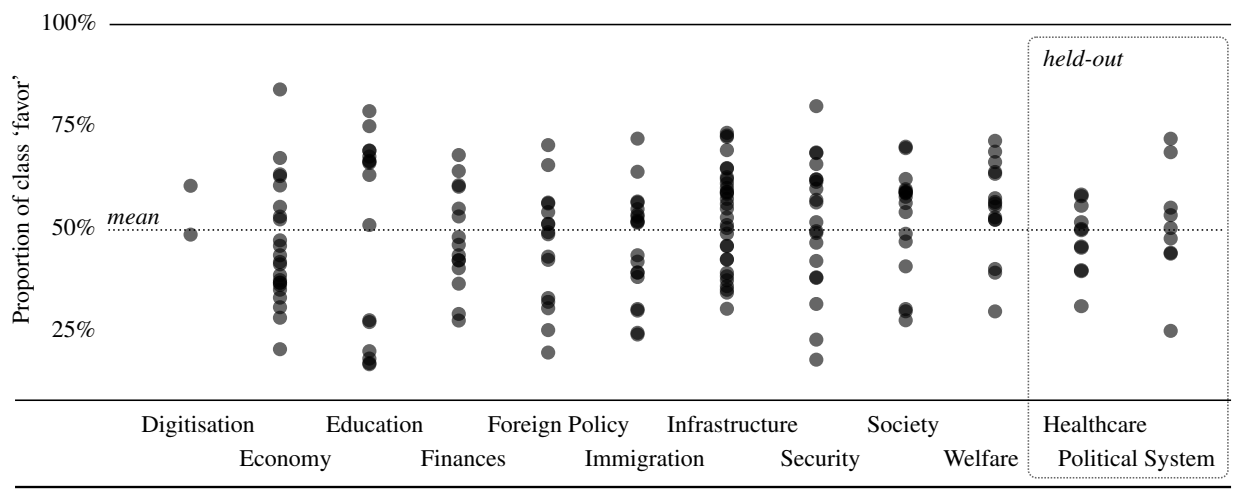}
\caption{
Proportion of `favor' labels per question, grouped by topic.
While the proportion of favorable answers varies from question to question, it is balanced overall.
}
\label{class_distributions}
\end{figure*}

We held out the topics ``Healthcare'' and ``Political System'' from the training data and created a separate \textit{cross-topic} test set that contains the questions and answers related to those topics.

Furthermore, in order to test \textit{cross-question} generalization performance within previously seen topics, we manually selected 16~held-out questions that are distributed over the remaining 10~topics.
We selected the held-out questions manually because we wanted to make sure that they are truly unseen and that no paraphrases of the questions are found in the training set.

We designated Italian as a test-only language, since relatively few comments have been written in Italian.
From the remaining German and French data we randomly selected a percentage of respondents as validation or as test respondents.

As a result we received one training set, one validation set and four test sets.
The sizes of the sets are listed in Table~\ref{split}.
We did not consider test sets that are cross-lingual and cross-target at the same time, as they would have been too small to yield significant results.

\subsection{Analysis}

Some observations regarding the composition of \xstance{} can be made.

\paragraph{Class Distribution}
Figure~\ref{class_distributions} visualizes the proportion of `favor' and `against` stances for each target in the dataset.
The ratio differs between questions but is relatively equally distributed across the topics.
In particular, the questions in the held-out topics (with a `favor' ratio of~49.4\%) have a similar class distribution as the questions in other topics (with a `favor' ratio of~50.0\%).

\paragraph{Linguistic Properties}
Not every question is unique; some questions are paraphrases describing the same political issue.
For example, in the 2015 election, the candidates were asked: \textit{``Should the consumption of cannabis as well as its possession for personal use be legalised?''} Four years later they were asked: \textit{``Should cannabis use be legalized?''} However, we do not see any need to consolidate those duplicates because they contribute to the diversity of the training data.

We further observe that while some questions in the dataset are quite short, some questions are rather convoluted.
For example, a typical long question reads:

\begin{quote}
    \small
    Some 1\% of direct payments to Swiss agriculture currently go to organic farming operations. Should this proportion be increased at the expense of standard farming operations as part of Switzerland's 2014-2017 agricultural policy?
\end{quote}
Such longer questions might be more challenging to process semantically.

\paragraph{Languages}
The \xstance{} dataset has more German samples than French samples.
The language ratio of about 3:1 is consistent across all training and test sets.
Given the two languages it is possible to either train two monolingual models or to train a single model in a multi-source setup~\cite{mcdonald-etal-2011-multi}.
We choose a multi-source baseline because \mbert{} is known to benefit from multilingual training data both in a supervised and in a cross-lingual scenario~\cite{kondratyuk-straka-2019-75}.

\section{Baseline Experiments}

We evaluate four baselines to obtain an impression of the difficulty of the task.

\subsection{Majority Class Baselines}
The first pair of baselines uses the most frequent class in the training set for prediction.
Specifically, the \textit{global} majority class baseline predicts the most frequent class across all training targets while the \textit{target-wise} majority class baseline predicts the class that is most frequent for a given target question.
The latter can only be applied to the intra-target test sets.

\subsection{Bag-of-Words Baseline}
As a second baseline, we train a \textit{fastText} bag-of-words linear classifier~\cite{joulin-etal-2017-bag}.
For each comment, we select the translation of the question that matches its language, and concatenate it to the comment.
We tokenize the text using the Europarl preprocessing tools~\cite{koehn2005europarl}.

The `against' class was slightly upsampled in the training data so that the classes are balanced when summing over all questions and topics.

We use the standard settings provided by the fastText library.\footnote{\url{https://github.com/facebookresearch/fastText}}
Optimal hyperparameters from the following range were determined based on the validation accuracy:
\begin{itemize}\itemsep0em
    \item Learning rate: 0.1, 0.2, 1
    \item Number of epochs: 5, 50
\end{itemize}
The word vectors were set to a size of 300.
We do not initialize them with pre-trained multilingual embeddings since preliminary experiments did not show a beneficial effect.

\subsection{Multilingual BERT Baseline}
As our main baseline model we fine-tune multilingual \bert{} (\mbert{}) on the task ~\cite{devlin-etal-2019-bert} which has been pre-trained jointly in 104 languages\footnote{\url{https://github.com/google-research/bert/blob/master/multilingual.md}} and has established itself as a state of the art for various multilingual tasks~\cite{wu-dredze-2019-beto, pires-etal-2019-multilingual}.
Within the field of stance detection, \bert{} can outperform both feature-based and other neural approaches in a monolingual English setting~\cite{ghosh2019stance}.

\paragraph{Architecture}
In the context of \bert{} we interpret the \xstance{} task as \textit{sequence pair classification} inspired by natural language inference tasks~\cite{bowman-etal-2015-large}.
We follow the procedure outlined by~\citet{devlin-etal-2019-bert} for such tasks.
We designate the question as segment \texttt{A} and the comment as segment \texttt{B}.
The two segments are separated with the special token \texttt{[SEP]}, and the special token \texttt{[CLS]} is prepended to the sequence.
The final hidden state corresponding to \texttt{[CLS]} is then classified by a linear layer.

We fine-tune the full model with a cross-entropy loss, using the AllenNLP library~\cite{gardner-etal-2018-allennlp} as a basis for our implementation.

\paragraph{Training}

As above, we balanced out the number of classes in the training set.
%
%
We use a batch size of~16 and a maximum sequence length of~512 subwords, and performed a grid search over the following hyperparameters based on the validation accuracy:
\begin{itemize}\itemsep0em
    \item Learning rate: 5e-5, 3e-5, 2e-5
    \item Number of epochs: 3, 4
\end{itemize}

The grid search was repeated independently for every variant that we test in the following subsections.
Furthermore, the standard recommendations for fine-tuning \bert{} were used:
Adam with ${\beta}_1=0.9$ and ${\beta}_2=0.999$; an L2 weight decay of~$0.01$; a learning rate warmup over the first~10\% of the steps; and a linear decay of the learning rate.
A dropout probability of~0.1 was set on all layers.

\paragraph{Results}

\begin{table}[]
\begin{tabularx}{\columnwidth}{@{}Xrr@{\hskip 18px}r@{}}
    \toprule
               & \textbf{\textsc{de}}      & \textbf{\textsc{fr}}     & \textbf{\textsc{it}}                         \\ \midrule
Majority class (global) & 33.1             & 34.8            & 34.4                                \\
Majority class (target-wise) & 60.8             & 65.1            & 59.3                                \\
fastText          & 69.9            & 71.2            & 53.7                                \\
\mbert{}          & 76.8            & 76.6            & 70.2                                \\ \bottomrule
\end{tabularx}
\caption{\label{cross-lingual-results}
Baseline scores in the \textit{cross-lingual} setting.
No Italian samples were seen during training, making this a case of zero-shot cross-lingual transfer.
The scores are reported as the macro-average of the F1-scores for `favor' and for `against'.
}
\end{table}

\begin{table*}[]
\begin{tabularx}{\textwidth}{@{}Xrrr@{\hskip 30px}rrr@{\hskip 30px}rrr@{}}
\toprule
               & \multicolumn{3}{l}{\phantom{q}\textbf{Intra-target}} & \multicolumn{3}{l}{\textbf{Cross-question}} & \multicolumn{3}{l}{\textbf{Cross-topic}} \\
               & \textsc{de}      & \textsc{fr}        & Mean     & \textsc{de}      & \textsc{fr}          & Mean       & \textsc{de}      & \textsc{fr}         & Mean      \\ \midrule
Majority class (global) & 33.1 & 34.8 & 33.9 & 36.4 & 37.9 & 37.1 & 32.1 & 33.8 & 32.9      \\
Majority class (target-wise) & 60.8 & 65.1 & 62.9 & - & - & - & - & - & -      \\
fastText       & 69.9 & 71.2 & 70.5 & 62.0 & 65.6 & 63.7 & 63.1 & 65.5 & 64.3      \\
\mbert{}       & 76.8 & 76.6 & 76.6 & 68.5 & 68.4 & 68.4 & 68.9 & 70.9 & 69.9      \\ \bottomrule
\end{tabularx}
\caption{\label{cross-target-results}
Baseline scores in the \textit{cross-target} setting.
For each test set we separately report a German and a French score, as well as their harmonic mean.
}
\end{table*}

Table~\ref{cross-lingual-results} shows the results for the cross-lingual setting.
\mbert{} performs consistently better than the previous baselines.
Even the zero-shot performance in Italian, while significantly lower than the supervised scores, is much better than the target-wise majority class baseline.

Results for the cross-target setting are given in Table~\ref{cross-target-results}.
Similar to the cross-lingual setting, model performance drops in the cross-target setting, but \mbert{} remains the strongest baseline and easily surpasses the majority class baselines.
Furthermore, the cross-question score of \mbert{} is slightly lower than the cross-topic score.

\subsection{How Important is Consistent Language?}

The default setup preserves \textit{horizontal language consistency} in that the language of the questions always corresponds to the language of the comments.
For example, the Italian test instances are combined with the Italian version of the questions, even though during training the model has only ever seen the German and French version of them.

An alternative concept is \textit{vertical language consistency}, whereby the questions are consistently presented in one language, regardless of the comment.
To test whether horizontal or vertical consistency is more helpful, we train and evaluate \mbert{} on a dataset variant where all questions are in their English version.
We chose English as a lingua franca because it had the largest share of data during the pre-training of \mbert{}.

Results are shown in Table~\ref{additional-results}.
While the effect is negligible in most settings, cross-lingual performance increases when all questions are in English.

\begin{table*}[]
\begin{tabularx}{\textwidth}{@{}Xrrrr@{}}
\toprule
                      & Supervised & Cross-Lingual & Cross-Question & Cross-Topic \\ \midrule
\mbert{}                 & \textbf{76.6}      &  70.2 & 68.4            & \textbf{69.9}                   \\
\textemdash{} with English questions   & 76.1       &\textbf{ 71.7}          & \textbf{68.5}         & 69.4        \\ \midrule
\textemdash{} with missing questions   & 73.2       & 67.1          & 67.8           & 69.3          \\
\textemdash{} with missing comments    & 64.2       & 60.5          & 51.1           & 48.6        \\
\textemdash{} with random questions    & 56.0       & 52.5            & 47.7             & 48.5           \\
\textemdash{} with random comments     & 50.7       & 50.7            & 48.2           & 48.7           \\ \midrule
\textemdash{} with target embeddings & 70.1      & 66.0          & 68.4          & 69.0     \\ \bottomrule
\end{tabularx}
\caption{\label{additional-results}
Results for additional experiments.
The cross-lingual score is the F1-score on the Italian test set.
For the supervised, cross-question and cross-topic settings we report the harmonic mean of the German and French scores.
}
\end{table*}

\subsection{How Important are the Segments?}

In order to rule out that only the questions or only the comments are necessary to optimally solve the task, we conduct some additional experiments:

\begin{itemize}
    \item Only use a single segment containing the comment, removing the questions from the training and test data (\textit{missing questions}).
    \item Only use the question and remove the comment (\textit{missing comments}).
\end{itemize}

In both cases the performance decreases across all evaluation settings (Table~\ref{additional-results}).
The loss in performance is much higher when comments are missing, indicating that the comments contain the most important information about stance.
As can be expected, the score achieved without comments is only slightly different from the target-wise majority class baseline.

But there is also a loss in performance when the questions are missing, which underlines the importance of pairing both pieces of text.
The effect of missing questions is especially strong in the supervised and cross-lingual settings.
To illustrate this, we provide in Table~\ref{ambiguous-comments} some examples of comments that occur with multiple different targets in the training set.
Those examples can explain why the target can be essential for disambiguating a stance detection problem.
On the other hand, the effect of omitting the questions is less pronounced in the cross-target settings.

\vphantom{}
 The above single-segment experiments tell us that both the comment and the question provide crucial information.
 But it is possible that the \mbert{} model, even though trained on both segments, mainly looks at a single segment at test time.
To rule this out, we probe the model with randomized data at test time:

\begin{itemize}
    \item  Test the model on versions of the test sets where the comments remain in place but the questions are shuffled randomly (\textit{random questions}).
    We make sure that the random questions come from the same test set and language as the original questions.
    \item Keep the questions in place and randomize the comments (\textit{random comments}).
    Again we shuffle the comments only within test set boundaries.
\end{itemize}

\noindent The results in Table \ref{additional-results} show that the performance of the model decreases in both cases, confirming that it learns to take into account both segments.

\subsection{How Important are Spelled-Out Targets?}

Finally we test whether the target really needs to be represented by natural language (e.g.~``Do~you support~X?'').
An alternative is to represent the target with a trainable embedding instead.

In order to fit target embeddings smoothly into our architecture, we represent each target type with a different reserved symbol from the \mbert{} vocabulary.
Segment \texttt{A} is then set to this symbol instead of a natural language question.

The results for this experiment are listed in the bottom row of Table~\ref{additional-results}.
An \mbert{} model that learns target embeddings instead of encoding a question performs clearly worse in the supervised and cross-lingual settings.
From this we conclude that spelled-out natural language questions provide important linguistic detail that can help in stance detection.

\section{Discussion}

Our experiments show that \mbert{} achieves a reasonable accuracy on \xstance{}, outperforming majority class baselines and a fastText classifier.

\begin{table}
\begin{tabularx}{\columnwidth}{@{}XXr@{}}
\toprule
\textbf{Dataset}  & \textbf{Evaluation}                         & \textbf{Score}     \\ \midrule
SemEval-2016  & \citet{ghosh2019stance}             & 75.1 \\
MPCHI    & \citet{ghosh2019stance}             & 75.6 \\
\xstance & this paper & 76.6   \\ \bottomrule
\end{tabularx}
\caption{\label{benchmark-comparison}
Performance of \bert{}-like models on different supervised stance detection benchmarks.
}
\end{table}

To put the supervised score into context we list scores that variants of \bert{} have achieved on other stance detection datasets in Table~\ref{benchmark-comparison}.
It seems that the supervised part of \xstance{} has a similar difficulty as the SemEval-2016~\cite{mohammad-etal-2016-dataset} or MPCHI~\cite{sen2018stance} datasets on which \bert{} has previously been evaluated.

On the other hand, in the cross-lingual and cross-target settings, the mean score drops by 6–8 percentage points compared to the supervised setting; while zero-shot transfer is possible to a degree, it can still be improved.

The additional experiments (Table~\ref{additional-results}) validate the results and show that the sequence-pair classification approach to stance detection is justified.

It is interesting to see what errors the \mbert{} model makes.
Table~\ref{errors} presents instances where it predicts the wrong label with a high confidence.
These examples indicate that many comments express their stance only on a very implicit level, and thus hint at a potential weakness of the dataset.
Because on the voting advice platform the label is explicitly shown to readers in addition to the comments, the comments do not need to express the stance explicitly.

Manual annotation could eliminate very implicit samples in a future version of the dataset.
However, the sheer size and breadth of the dataset could not realistically be achieved with manual annotation, and, in our view, largely compensates for the implicitness of the texts.

\section{Conclusion}

We have presented a new dataset for political stance detection called \xstance{}.
The dataset extends over a broad range of topics and issues regarding national Swiss politics.
This diversity of topics opens up an opportunity to further study multi-target learning.
Moreover, being partly Swiss Standard German, partly French and Italian, the dataset promotes a multilingual approach to stance detection.

By compiling formal commentary by politicians on political questions, we add a new text genre to the field of stance detection.
We also propose a question--answer format that allows us to condition stance detection models on a target naturally.

Our baseline results with multilingual \bert{} show that the model has some capability to perform zero-shot transfer to unseen languages and to unseen targets (both within a topic and to unseen topics).
However, there is some gap in performance that future work could address.
We expect that the \xstance{} dataset could furthermore be a valuable resource for fields such as argument mining, argument search or topic classification.

\section*{Acknowledgments}
This work was funded by the Swiss National Science Foundation (project MUTAMUR; no.~176727).
We would like to thank Isabelle \mbox{Augenstein}, Anne Göhring and the anonymous reviewers for helpful feedback.

\bibliography{acl2019}
\bibliographystyle{acl_natbib}

\appendix

\clearpage
\onecolumn

\section{Examples}

\renewcommand{\thetable}{\Alph{section}\arabic{table}}

\begin{table*}[!htb]
    \small
\begin{tabularx}{\textwidth}{@{}XXlr@{}}
\toprule
\textbf{Question}                                                                                                                        & \textbf{Comment}                                                                                                                  & \textbf{Gold Label} & \textbf{Prob.} \\ \midrule
\textit{Befürworten Sie eine vollständige Liberalisierung der Geschäftsöffnungszeiten?} &
\textit{Ausser Sonntag. Dies sollte ein Ruhetag bleiben können.}
& FAVOR      & 0.001                               \\
{[}Are you in favour of a complete liberalisation of business hours for shops?{]} &
{[}Except Sunday. That should remain a day of rest.{]}
& & \\ \midrule
\textit{Soll die Schweiz innerhalb der nächsten vier Jahre EU-Beitrittsverhandlungen aufnehmen?} &
\textit{In den nächsten vier Jahren ist dies wohl un\-realistisch.}
& FAVOR      & 0.005    \\
{[}Should Switzerland embark on negotiations in the next four years to join the EU?{]} &
{[}For the next four years this is probably unrealistic.{]}
& & \\ \midrule
\textit{Befürworten Sie einen Ausbau des Landschaftsschutzes?} &
\textit{Wenn es darum geht erneuerbare Energien zu fördern, ist sogar eine Lockerung angebracht.}
& AGAINST      & 0.006    \\
{[}Are you in favour of extending landscape protection?{]} &
{[}When it comes to promoting renewable energy, even a relaxation is appropriate.{]}
& & \\ \midrule
\textit{La Suisse devrait-elle engager des négociations pour un accord de libre échange avec les Etats-Unis?} &
\textit{Il faut cependant en parallèle veiller à ce que la Suisse ne soit pas mise de côté par les Etats-Unis~!}
& AGAINST      & 0.010    \\
{[}Should Switzerland start negotiations with the USA on a free trade agreement?{]} &
{[}At the same time it must be ensured that Switzerland is not sidelined by the United States!{]}
& & \\ \bottomrule
\end{tabularx}
\caption{\label{errors}
Some classification errors where the predicted probability of the correct label is especially low.
The examples have been taken from the validation set.
}
\end{table*}

\vphantom{}
\vphantom{}
\vphantom{}

\begin{table*}[!htb]
    \small
\begin{tabularx}{\textwidth}{@{}XXX@{}}

\toprule
\textbf{Comment} \ldots                                                                                                                  & \textbf{is favorable towards target \ldots}                                                                                   & \textbf{but against target \ldots}                                                                           \\ \midrule
\textit{Ich will offene Grenzen für Waren und selbstverantwortliche mündige Bürger. Der Staat hat kein Recht, uns einzuschränken.} & \textit{Soll die Schweiz mit den USA Verhandlungen über ein Freihandelsabkommen aufnehmen?}
& \textit{Soll die Schweiz das Schengen-Abkommen mit der EU kündigen und wieder verstärkte Personenkontrollen direkt an der Grenze einführen?} \\
{[}I want open borders for goods and responsible citizens. The state has no right to restrict us.{]} & {[}Should Switzerland start negotiations with the USA on a free trade agreement?{]} & {[}Should Switzerland terminate the Schengen Agreement with the EU and reintroduce increased identity checks directly on the border?{]} \\ \midrule
\textit{Hier gilt der Grundsatz der Eigenverantwortung und Selbstbestimmung des Unternehmens!}                                     & \textit{Sind Sie für eine vollständige Liberalisierung der Ladenöffnungszeiten?} & \textit{Würden Sie die Einführung einer Frauenquote in Verwaltungsräten börsenkotierter Unternehmen befürworten?}    \\
{[}The principle of personal responsibility and corporate self-regulation applies here!{]} & {[}Are you in favour of the complete liberalization of shop opening times?{]} & {[}Would you support the introduction of a woman's quota for the Boards of Directors of listed companies?{]}   \\ \bottomrule
\end{tabularx}
    \caption{\label{ambiguous-comments}
Two comments that imply a positive stance towards one target issue but a negative stance towards another target issue. Such cases can be found in the dataset because respondents have copy-pasted some comments.
These examples have been extracted from the training set.
}
\end{table*}

\clearpage

\section{Data Statement}
\label{appendix:data-statement}

\paragraph{Curation rationale}
In order to study the automatic detection of stances on political issues, questions and candidate responses on the voting advice application \texttt{smartvote.ch} were downloaded.
Mainly data pertaining to national-level issues were included to reduce variability.

\paragraph{Language variety}
The training set consists of questions and answers in Swiss Standard German and Swiss French (74.1\% de-CH; 25.9\% fr-CH).
The test sets also contain questions and answers in Swiss Italian (67.1\% de-CH; 24.7\% fr-CH; 8.2\% it-CH).
The questions have also been translated into English.

\paragraph{Speaker demographic (answers)}
\begin{itemize}
    \item Candidates for communal, cantonal or national elections in Switzerland who have filled out an online questionnaire.
    \item Age: 18 or older -- mixed.
    \item Gender: Unknown -- mixed.
    \item Race/ethnicity: Unknown -- mixed.
    \item Native language: Unknown -- mixed.
    \item Socioeconomic status: Unknown -- mixed.
    \item Different speakers represented: 7581.
    \item Presence of disordered speech: Unknown.
\end{itemize}


\paragraph{Speech situation}
\begin{itemize}
  \item The questions were edited and translated by political scientists for a public voting advice website.
  \item The answers were written between 2011 and 2020 by the users of the website.
\end{itemize}

\paragraph{Text characteristics} Questions, answers, arguments and comments regarding political issues.

\end{document}